%% file: main.tex
\documentclass[11pt]{article}

\usepackage[preprint]{acl}

\usepackage{times}
\usepackage{latexsym}

\usepackage[T1]{fontenc}
\usepackage[utf8]{inputenc}

\usepackage{microtype}

\usepackage{inconsolata}

\usepackage{graphicx}

\usepackage{amsmath}
\usepackage{booktabs}
\usepackage{multirow}
\usepackage{amsfonts}
\usepackage{amssymb}
\usepackage{tabularx}

\title{An Exploratory Study to Repurpose LLMs to a Unified Architecture for Time Series Classification}

\author{Hansen He \\
  Canyon Crest Academy \\
  \texttt{hansenhe2008@gmail.com} \\\And
 Shuheng Li \\
  UC San Diego \\
  \texttt{shl060@ucsd.edu}}

\begin{document}
\maketitle
\begin{abstract}

\input{0-abstract}

\end{abstract}

\input{1-introduction}

\input{2-related}

\input{3-method}

\input{4-evaluation}

\input{5-conclusion}

\bibliography{ref}

% \appendix

% \section{Example Appendix}
% \label{sec:appendix}

% This is an appendix.

\end{document}

%% file: 0-abstract.tex
Time series classification (TSC) is a core machine learning problem with broad applications. Recently there has been growing interest in repurposing large language models (LLMs) for TSC, motivated by their strong reasoning and generalization ability. Prior work has primarily focused on alignment strategies that explicitly map time series data into the textual domain; however, the choice of time series encoder architecture remains underexplored. In this work, we conduct an exploratory study of hybrid architectures that combine specialized time series encoders with a frozen LLM backbone.
%Raw time series are first transformed into compact latent representations using classical encoders, which are then injected into the LLM via its input embedding layer for classification.
We evaluate a diverse set of encoder families, including Inception, convolutional, residual, transformer-based, and multilayer perceptron architectures, among which the Inception model is the only encoder architecture that consistently yields positive performance gains when integrated with an LLM backbone. Overall, this study highlights the impact of time series encoder choice in hybrid LLM architectures and points to Inception-based models as a promising direction for future LLM-driven time series learning.

%% file: 1-introduction.tex
\section{Introduction}
\label{sec:intro}
TSC is a fundamental machine learning task with applications across a wide range of domains, including healthcare, finance, and smart building systems. Over the past decade, advances in deep neural networks (NNs) have led to state-of-the-art performance in TSC using classical architectures such as convolutional neural networks (CNNs) and recurrent neural networks (RNNs) \cite{ismail2019deep}. However, these approaches are typically tailored to specific application domains, relying on careful architectural selection and extensive hyperparameter tuning to achieve optimal performance. This domain-specific design arises from the heterogeneous nature of TSC tasks, which often require distinct feature representations. For instance, in healthcare applications, classifying electrocardiogram (ECG) signals for cardiac abnormality detection depends on capturing fine-grained morphological patterns in heartbeat waveforms, whereas in sensing systems with high-frequency sampling, frequency-domain characteristics are critical for tasks such as activity recognition from accelerometer data.

Recently, LLMs have emerged as a general-purpose solution across diverse problem domains due to their task-agnostic architectures and pretraining on large-scale datasets that enable rich representation learning. This progress motivates the exploration of repurposing LLMs as a unified framework for TSC. A straightforward approach treats numerical time series values as textual inputs and applies prompt-based formatting \cite{xue2023promptcast}. However, LLMs are fundamentally designed to operate on discrete text tokens rather than continuous numerical signals. This mismatch introduces quantization effects during tokenization, leading to precision loss and potentially obscuring fine-grained temporal patterns that are critical for accurate classification.

To address these limitations, recent studies have proposed hybrid architectures that combine specialized time series encoders with LLM backbones. Existing work primarily focuses on designing representation alignment frameworks to map time series data into the textual domain \cite{gruver2023large, kaur2024lets}.

Despite their demonstrated effectiveness, the role of time series encoder architecture in such hybrid systems remains relatively underexplored. In this paper, we aim to fill this gap through a systematic experimental study of encoder design choices for LLM-based TSC. Using the 2015 UCR Time Series Archive \cite{UCRArchive}, we evaluate a range of encoder families—including MLP, CNN, Inception, ResNet, and transformer-based architectures—when stacked with a frozen LLM, as well as their standalone counterparts. Our results show that incorporating an LLM generally degrades classification performance, with the notable exception of the Inception-based encoder, which not only benefits from LLM integration but also achieves the highest standalone accuracy. We hypothesize that this advantage stems from the Inception architecture’s ability to capture multi-scale temporal patterns through parallel convolutional kernels of varying receptive fields, while remaining sufficiently lightweight for small-scale time series datasets.

In summary, this work presents an exploratory investigation into hybrid LLM-based architectures for time series classification, offering empirical insights into the impact of encoder choice and establishing a foundation for future research on LLM-driven time series analysis.

%% file: 2-related.tex
\section{Related Works}

Deep learning-based approaches have demonstrated strong performance on TSC tasks. Early studies adapted classical neural architectures originally developed for natural language processing and computer vision to address TSC problems, achieving competitive results~\cite{zhao2017convolutional,malhotra2017timenet,du2018attention}. To further improve performance and mitigate domain-specific discrepancies, subsequent research has proposed models tailored to particular time series characteristics or application domains. For instance, Medformer~\cite{wang2024medformer} introduces a cross-channel patching mechanism designed specifically for multivariate medical time series. STFNets~\cite{yao2019stfnets} integrates multi-resolution Short-Time Fourier Transform representations into deep neural networks to capture frequency-domain features in sensing time series data. While these domain-specific methods achieve state-of-the-art performance within their respective settings, their generalize-ability across diverse TSC domains remains largely unverified.

Motivated by the strong reasoning capabilities of LLMs, recent work has explored using LLMs as an unified solution to time series analysis. Xue et al.~\cite{xue2023promptcast} and Gruver et al.~\cite{gruver2023large} treat time series values as text tokens, leveraging LLMs’ next-token prediction capability for time series forecasting. However, since the textual representation of numerical values does not inherently capture their underlying mathematical relationships, subsequent studies have focused on lightweight fine-tuning or architectural adaptations of LLMs for TSC. Zhou et al.~\cite{zhou2023one} propose a patch-based encoder to extract time series representations suitable for language models. LETS-C~\cite{kaur2024lets} employs a pre-trained language embedding model to encode raw time series data, followed by a CNN–MLP classifier for downstream classification. Tao et al.~\cite{tao2024hierarchical} further propose explicitly aligning representations in the semantic space of language models with those in the time series feature space.

Although these approaches provide promising solutions for bridging the gap between time series and textual representations, the optimal architectural design for encoding time series data, particularly in the context of LLM-based models, remains an open research question. This paper aims to address this gap by systematically investigating effective architectures for time series encoding.

%% file: 3-method.tex
\section{Methodology}

\subsection{Problem Definition}

Let $\mathcal{D} = \{(X^{(i)}, y^{(i)})\}_{i=1}^N$ be a dataset containing $N$ labeled time series samples, where each $X^{(i)} = [x^{(i)}_1, x^{(i)}_2, \dots, x^{(i)}_T] \in \mathbb{R}^{T \times d}$ represents a sequence of time series data in a window size of $T$ and $y^{(i)} \in \{1, 2, \dots, C\}$ is the class label among $C$ predefined categories. The objective of TSC is to learn a mapping $f: \mathbb{R}^{T \times d} \rightarrow \{1, 2, \dots, C\}$ parameterized by $\theta$, such that the predicted label $\hat{y} = f(X)$
matches the ground truth label $y$ with high accuracy.
In deep learning-based approaches, the mapping function $f_\theta(\cdot)$ is optimized over the training dataset $\mathcal{D}$. However, when the data distribution or task specification changes, traditional TSC models typically require redesigning the model architecture and retraining from scratch, resulting in substantial computational and engineering overhead.

Motivated by the strong generalization and reasoning capabilities of LLMs, this work investigates effective time series encoder architectures that can be stacked on top of LLMs, enabling reusable and flexible representations for TSC.

\subsection{Time Series Encoder}
LLMs are inherently designed to process sequences of discrete tokens rather than continuous-valued time series. A straightforward approach is to convert numerical time series into textual descriptions and feed them directly into an LLM. However, such tokenization often incurs information loss, introduces inefficiencies, and provides a weak inductive bias for modeling temporal dynamics, as discussed in Section~\ref{sec:intro}.

To overcome these limitations, we propose to first encode raw time series into compact latent representations using a dedicated time series encoder, and subsequently combine these representations with an LLM. This design allows temporal feature extraction to be handled by specialized encoders, while the LLM contributes high-level semantic reasoning.

\paragraph{Encoder Architecture.} 

Formally, let $E_\phi: \mathbb{R}^{T \times d} \rightarrow \mathbb{R}^{L \times h}$ denote a time series encoder parameterized by $\phi$, where $L$ is the number of encoded latent tokens and $h$ matches the hidden dimensionality of the downstream LLM. The encoder $E_\phi$ transforms the input time series into a sequence of latent vectors that summarize temporal patterns and dependencies in an LLM-compatible format.

In this work, we systematically evaluate multiple encoder architectures, including MLP, CNN, Inception-style networks, and Transformer-based encoders, in order to assess their effectiveness for downstream TSC when coupled with LLMs.

% The encoder outputs
% \begin{equation}
% Z = E_\phi(X) = [z_1, z_2, \dots, z_L], \quad z_i \in \mathbb{R}^h,
% \end{equation}
% which captures temporal dynamics and local patterns in a way that is information-preserving yet LLM-friendly.

\paragraph{Motivation for Encoder-Layer Stacking.} 
Encoding continuous time series into learned latent representations prior to LLM processing offers several advantages:
\begin{itemize}
\item It preserves fine-grained temporal structure without relying on textual tokenization.
\item It significantly reduces sequence length, improving computational efficiency and enabling the LLM to focus on higher-level reasoning.
\item It facilitates cross-modal integration, allowing the LLM to serve as a semantic reasoning head atop temporal feature representations.
\end{itemize}

Given the encoded representation $Z = E_\phi(X)$, we concatenate $Z$ with prompt and padding embeddings to form the LLM input $\hat{Z}$. The final prediction is obtained as $\hat{y} = \mathrm{LLM}_\psi(\hat{Z})$, where $\psi$ denotes the frozen parameters of the LLM. The overall model is trained end-to-end with respect to the encoder parameters $\phi$ and the classification head, while keeping the LLM fixed.

%% file: 4-evaluation.tex
\section{Evaluation}

\subsection{Experimental Setup}

%% Dataset
We evaluate our framework on the 2015 UCR Time Series Archive~\cite{UCRArchive}, a benchmark time-series datasets which spans diverse domains such as sensor, motion, and physiological data. Each dataset follows standard train/test splits.

%% Model, baseline and variants

We evaluate three architectural categories implemented in our codebase\footnote{https://github.com/rwt-9829/tsc-llm}:
\begin{itemize}
    \item \textbf{TS\_CLS\_NoEncoder}: Llama-3.1-8B used directly for zero/few-shot classification via textual prompts without a time-series encoder.
    \item \textbf{TS\_CLS\_NoLLM}: Traditional encoder-only models including MLP, CNN, ResNet and Transformer followed by a linear prediction head.
    \item \textbf{TS\_CLS}: Hybrid models combining each encoder with a frozen Llama-3.1-8B backbone. 
\end{itemize}

%%  Detailed setup, including hyperparameters (adam, epochs, learning rate etc.) Other details about the model architecture (that is not covered in the method section)
All models were trained using the ADAM optimizer to 100 epochs with a learning rate of $0.001$ and batch size of $32$ unless otherwise specified. The experiments are conducted using NVIDIA L40S GPUs with CUDA Version 12.4.
%% Implementation details (including coding language, package, gpu, server etc.)
% All experiments are implemented in PyTorch. Llama-3.1-8B is loaded using the \texttt{transformers} library (\texttt{AutoModel} and \texttt{AutoTokenizer}). The LLM is frozen during training and receives encoder-generated embeddings through its input embedding layer.
%% TODO GPU AND SERVER INFORMATION

\subsection{Main Results}

Table~\ref{tab:main_res} reports the average accuracy across all datasets for each evaluated architecture, with and without LLM integration. Overall, the Inception-based encoder achieves the highest performance among the compared architectures and is the only model that shows a consistent accuracy improvement when combined with an LLM. We hypothesize that this effect arises from the use of multi-scale convolutional kernels in the Inception architecture, which provide receptive fields at multiple resolutions. Such a design may enable more effective modeling of time-series patterns across different temporal scales, potentially leading to better generalization than the other architectures.

\begin{table}[t]
\centering
\caption{Comparison of Model Architectures}
\begin{tabular}{lcccccc}
\toprule
\multirow{2}{*}{Model} & \multicolumn{2}{c}{Max Test Acc.} & \multicolumn{2}{c}{Min Loss Acc.} \\
\cmidrule(r){2-3} \cmidrule(r){4-5}
 & Plain & + LLM & Plain & + LLM \\
\midrule
\textbf{Inception}         & 71.15 & 74.21 & 67.07 & 69.34 \\
\textbf{MLP}               & 62.42 & 56.75 & 59.59 & 52.28 \\
\textbf{Transfomer}         & 42.63 & 42.24 & 39.12 & 34.94 \\
\textbf{CNN}               & 68.59 & 62.25 & 57.53 & 47.32 \\
\textbf{ResNet}            & 70.74 & 69.59 & 64.66 & 60.09 \\
\bottomrule
\end{tabular}
\label{tab:main_res}
\end{table}

\subsection{Hyperparameter Tuning}

We conduct a coarse grid search over a small set of core hyperparameters to further identify the best setting. Specifically, we vary the learning rate over $\{10^{-3}, 10^{-4}, 10^{-5}\}$, the number of kernels $N_{\text{kernels}}$ over $\{3, 4, 5, 6\}$, and the kernel size parameter $K$ over $\{8, 16\}$. Each configuration is evaluated on a subset of 10 representative datasets drawn from the entire set. Selected testing result is shown in Table~\ref{tab:grid_search}. We observe that a larger number of kernels and kernel sizes generally leads to better performance, which supports our conjecture that varying reception fields contribute to TSC accuracy.

% Each configuration is evaluated on a subset of 10 representative datasets drawn from the benchmark: \texttt{uWaveGestureLibrary\_Y}, \texttt{Cricket\_Z}, \texttt{Beef}, \texttt{InlineSkate}, \texttt{Ham}, \texttt{FordA}, \texttt{DistalPhalanxOutlineCorrect}, \texttt{Cricket\_Y}, \texttt{ToeSegmentation1}, and \texttt{Computers}. Performance is measured using \textit{average maximum test accuracy}.

\begin{table}[t]
\centering
\caption{Selected Grid search of learning rate, number of kernels $N_{\text{kernels}}$, and kernel size $K$ over subset of $10$ datasets}
\begin{tabular}{lccc}
\toprule
Learning Rate & $N_{\text{kernels}}$ & $K$ & Avg. Max Acc. \\
\midrule
$10^{-3}$ & 3 & 8  & 0.6485 \\
$10^{-3}$ & 3 & 16 & 0.6454 \\
$10^{-3}$ & 4 & 8  & 0.6330 \\
$10^{-3}$ & 4 & 16 & 0.6506 \\
$10^{-3}$ & 5 & 8  & 0.6488 \\
$10^{-3}$ & 5 & 16 & \textbf{0.6568} \\
$10^{-3}$ & 6 & 8  & \textbf{0.6552} \\
$10^{-3}$ & 6 & 16 & \textbf{0.6552} \\
\midrule
% $10^{-4}$ & 3 & 8  & 0.6107 \\
% $10^{-4}$ & 3 & 16 & 0.6294 \\
% $10^{-4}$ & 4 & 8  & 0.6238 \\
% $10^{-4}$ & 4 & 16 & 0.6346 \\
% $10^{-4}$ & 5 & 8  & 0.6448 \\
$10^{-4}$ & 5 & 16 & 0.6103 \\
$10^{-4}$ & 6 & 8  & 0.6357 \\
$10^{-4}$ & 6 & 16 & 0.6322 \\
\midrule
% $10^{-5}$ & 3 & 8  & 0.4407 \\
% $10^{-5}$ & 3 & 16 & 0.5065 \\
% $10^{-5}$ & 4 & 8  & 0.4897 \\
% $10^{-5}$ & 4 & 16 & 0.5122 \\
% $10^{-5}$ & 5 & 8  & 0.4910 \\
$10^{-5}$ & 5 & 16 & 0.4866 \\
$10^{-5}$ & 6 & 8  & 0.5002 \\
$10^{-5}$ & 6 & 16 & 0.4869 \\
\bottomrule
\end{tabular}
\label{tab:grid_search}
\end{table}

%% file: 5-conclusion.tex
\section{Conclusion}

We examined whether LLMs can be repurposed for TSC by stacking them with various encoder families. 
In summary, experiments on the 2015 UCR Time Series Archive~\cite{UCRArchive} demonstrate that encoder architecture is a key factor in hybrid LLM-based TSC, with only the Inception-based encoder consistently yielding performance gains when combined with an LLM backbone, likely due to its effective multi-scale feature extraction across diverse domains.